\documentclass[10pt,conference]{IEEEtran}
\usepackage{geometry}
\usepackage{amsmath} 
\usepackage{algorithm}  
\usepackage{algpseudocode,algorithmicx}
\usepackage{slashed} 
\usepackage{wasysym}
\usepackage{bm}
\usepackage{mathrsfs}
\usepackage{graphicx}
\usepackage{subfigure}

\usepackage{cite}
\ifCLASSINFOpdf
\else
\fi
\usepackage{amsmath}
\usepackage{array}
\usepackage{stfloats}

\begin{document}

\title{Deformable Deep Convolutional Generative Adversarial Network in Microwave Based Hand Gesture Recognition System}

%\author{\IEEEauthorblockN{Jiajun Zhang\IEEEauthorrefmark{1},
%Jinkun Tao\IEEEauthorrefmark{1},
%James Kirk\IEEEauthorrefmark{1},
%Montgomery Scott\IEEEauthorrefmark{1} and
%Eldon Tyrell\IEEEauthorrefmark{1}
%}
%\IEEEauthorblockA{\IEEEauthorrefmark{1}College of Information Science \& Electronic Engineering\\
%Zhejiang University, Hangzhou, China \\ Email: justinzhang1993@gmail.com}
%}

\author{Jiajun Zhang, Zhiguo Shi \\
	\IEEEauthorblockA{
		College of Information Science \& Electronic Engineering\\
		Zhejiang University, Hangzhou, China\\
		{\{justinzhang,~shizg\}@zju.edu.cn}
	}\\
	%\vspace{-0.5cm}
	%\authorblockA{~~}
}

\maketitle

\begin{abstract}
Traditional vision-based hand gesture recognition systems is limited under dark circumstances. In this paper, we build a hand gesture recognition system based on microwave transceiver and deep learning algorithm. A Doppler radar sensor with dual receiving channels at 5.8GHz is used to acquire a big database of hand gestures signals. The received hand gesture signals are then processed with time-frequency analysis.  Based on these big database of hand gesture, we propose a new classification architecture called deformable deep convolutional generative adversarial network. Experimental results shows the new architecture can upgrade the recognition rate by 10\% and the deformable kernel can reduce the testing time cost by 30\%.
\end{abstract}

\IEEEpeerreviewmaketitle

\section{Introduction}
Hand gesture recognition is always a popular study topic in the computer science, due to its wide application in real life  systems like security safety and entertainment devices. The traditional hand gesture recognition systems basically relies on cameras and videos\cite{hjelmaas2001face}.Though camera-based recognition techniques are mature, they have some inborn limitations Firstly, they are not work well in the dark situation because the system is based on visual and recognition rate will intensely fall\cite{zhao2003face}. Besides, the system requires high computational and power resources, which restrained them from standalone usage. In addition, the nature of visual-based recognition system will cause the potential privacy leakage problems.

Recently, hand gesture recognition based on microwave has begun to gain popularity in public view\cite{molchanov2015multi,molchanov2015short,arbabian201394}. Compared with traditional methods, microwave-based hand gesture recognition systems has its own advantages. To begin with, while camera sets can severely affected by the light condition, microwave signals are not free of lightings and can even transmitting through total darkness. Secondly, microwave signal can generated and received by cost-effective but well designed microchips. Thus, it can be energy-friendly and independent from expensive hardware, which enables wide application of this technique on standalone devices possible. Thirdly, the microwave-based system obviously has no worries about privacy leakage. Because microwave-based on process the signal but not the images.

However, compared to the massive number of researches on the visual-based hand gesture recognition system, there are few literatures about the radar's usage on gesture recognition until recently. Some of the researches only focus on radar's signal on E-band or WLAN band. In \cite{hugler2016rcs}, the author use the mono-static radar cross section measurements of a human hand for radar-based gesture recognition system in E-band. The obvious limitation of this certain system is E-band is too high and too expensive for wide applications on standalone devices in real life. In addition, the researcher in \cite{pu2013whole} proposed a gesture recognition system that adopt Wi-Fi signals to enable whole-home sense network. Because Wi-Fi signal can travel through walls. However, 2.4GHz Wi-Fi signal is too crowded for massive application. On the other hand, there are several hand gesture recognition works focus on human falling detection. High accuracy falling detection from normal movements was achieved by Zig-bee module in computer\cite{mercuri2013analysis}. In addition, researchers adopt ultra-band microwave\cite{zhou2016ultra} and coherent frequency-modulated continuous-wave(FMCW)\cite{peng2016fmcw} in their system. However, these attempts focus mainly on motion detection rather than fine-grained signal processing.

Based on the good quality of microwaves, we attempt to build a hand gesture recognition system based on microwave radar and try to solve the problem using deep learning algorithm on big database. To get the big database of hand gesture, we firstly have to build a hardware architecture. In our system, we use a Doppler-radar sensor with dual receiving channels at 5.8GHz to acquire of massive samples of 24 standard hand gesture combination categories. Then we applied two time-frequency analysis, short-time Fourier transform and continuous wavelet transform, to the time-domain signal, and use those results as the foundation of the classification. Here comes the tricky part. When we do the classification using traditional convolutional neural network, it works really well when the standard gesture category is small. But the over-fitting problem of the model appears when the number of categories became larger. This is due to the reality that we have limited samples to train the model. Another problem we are facing is that the testing time on the demo became too longer when size of the sample database increases and the model become too complex. So, the main focus of our work is to tackle two problems rising with the expanding of gestures and limited database:the over fitting performance and the long testing time. 

The main contribution of this paper is that we propose a Deformable Deep Convolutional Generated Adversarial Network (De-DCGAN) as the classification method to solve these two problems. 
\begin{enumerate}
	\item Based on the GAN(Generated Adversarial Networks), we replace pooling layers with strides convolutions in discriminator and fractional-strided convolutions in generators.
	\item We adopt scaled exponential linear units as activation layers in generators and discriminators, which enable the network to self-normalize.
	\item We utilize deformable convolutional kernel in all the convolutional layers.
\end{enumerate}

After training on the real hand gesture database we collect, we can use the discriminator individually as the classifier. Several experiments are performed to test the modifications' performance on our database. The experiments shows that the proposed architecture not only increase the correct recognition rate with different gesture combinations, but also decrease the time of testing. To sum up, our proposed system proves the feasibility of the De-DCGAN of tackling time and over-fitting problems. And the experiments shows a great potential on De-DCGAN's performance on microwave-based gesture recognition system.

The paper is structured as follows. Section II gives the system design and the time-frequency analysis. Section III introduces the deformable deep convolutional generative adversarial network. Section IV discusses the results of our experiments. Finally, conclusion remarks are drawn in Section V.

\section{System Design and Pre-analysis}

\subsection{System outline}
\begin{figure}[htb]
	\centerline{\includegraphics[height=5cm,clip]{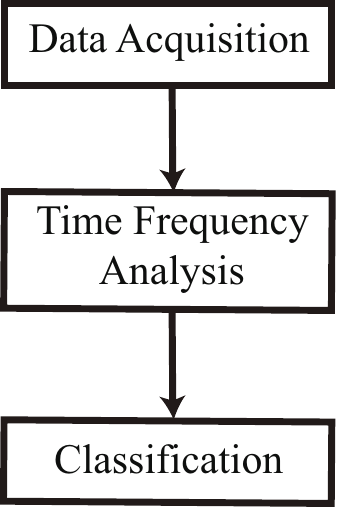}}
	\caption{System Operation Flow}
	\label{outline} 		
\end{figure}

The operation flow of hand gesture recognition system is approved as Fig. \ref{outline}. The first step is data acquisition. The database is essentially composed of the received signal samples from different hand gesture. We establish 24 standard gestures (combinations from 4 basic gestures circle, square, tick, cross) as examples in our research.

Then, we do time-frequency analysis of the time-domain signals from the database. Here, we use two stable and efficient time-frequency analysis algorithms. The first is known as short-time Fourier transform and the second is called continuous wavelet transform. The analysis results are the foundations of the classification.

Finally, we proposed a new Deformable Deep Convolutional Generative Adversarial Network(De-DCGAN) architecture derived from Deformable Convolutional Network(DCN) and Deep Convolutional Generated Adversarial Network(DCGAN) as the classification method. In this step,we mainly focus on two problems, the first is the over-fitting performance and the second is the long testing time. Based on the DCGAN architecture, we made a modifications to original DCGAN and  replace the kernel with deformable ones. Following are the specific design of each processing step in the system.

\subsection{Data Acquisition}

In order to obtain a well-performance and cost-effective transceiver platform for applications, we design a coherent zero Intermediate Frequency (IF) architecture with a symmetric sub-carrier modulation.

The system is designed to operate in the 5.8GHz band. Two different transmit and receive channels are designed to be based on individual modules for different chips for ordinary Wi-Fi applications. By the bandwidth of the chip, the frequency of the two sub-carriers will be set to about 6MHz. Select ADI's AD7357 two ADC chips for bandpass sampling. The sampling frequency used is about 100 Hz, well below all sub-carriers Finally, the USB port will receive the sampled data from the computer acquisition port.

\begin{figure}[htb]
	\centerline{\includegraphics[height=5cm,clip]{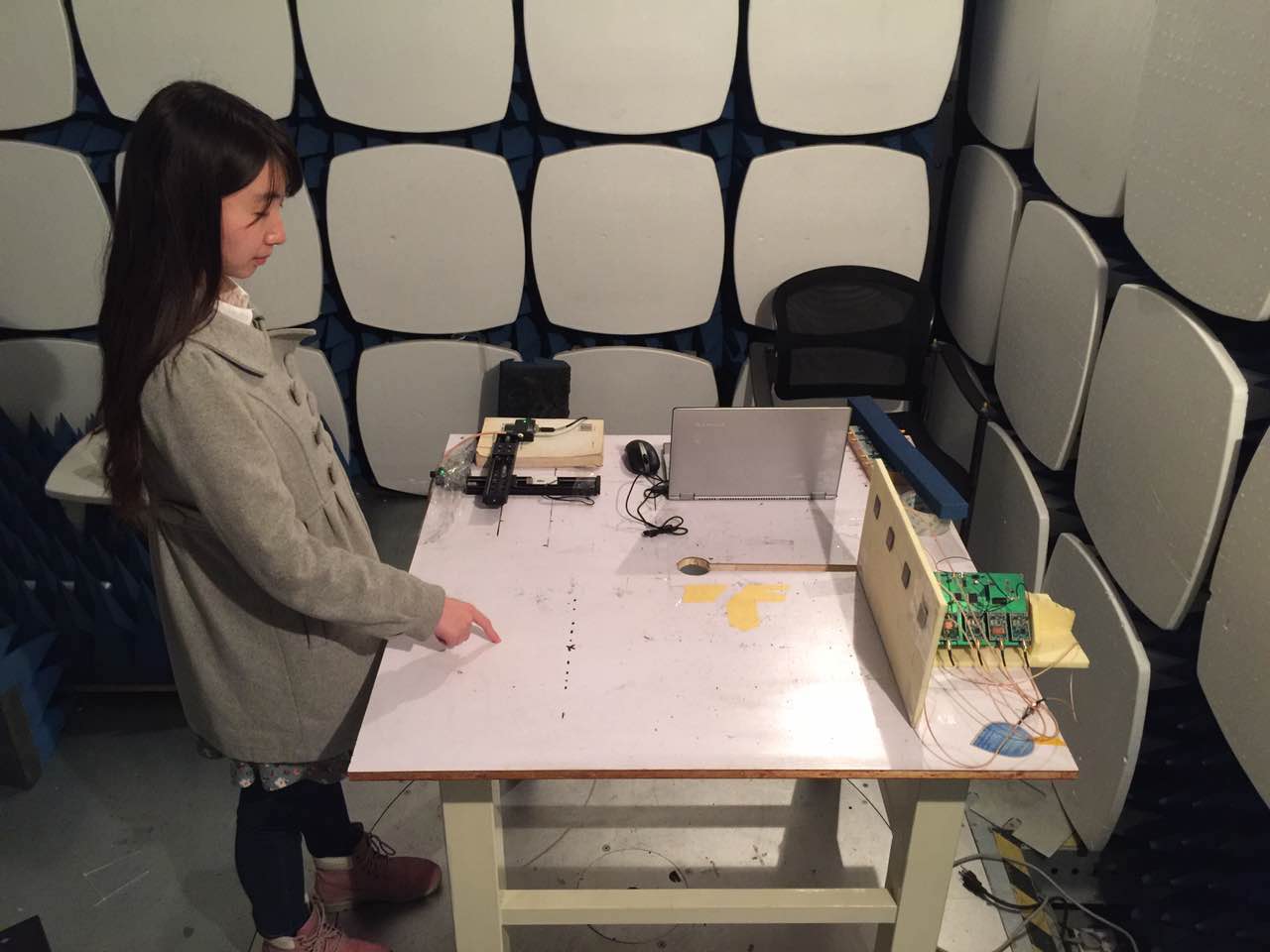}}
	\caption{Experiment setup for data acquisition}
	\label{posing}
\end{figure}

Fig. \ref{posing} shows the completed experiment setup of how the data is collected. In the conducted experiments, this setting is placed in a conventional laboratory environment with thick walls and metal cabinets. In our hand gesture recognition tranceiving system, 24 hand gestures are set as the basic reference hand gestures, which are combinations of circle, square, right, cross. All the gestures are captured at a same speed and the same time window of 1 second. And the gestures are in the horizontal plane in front of the antennas. At last, a total of 38400 samples were acquired at different distances from the tranceiving antennas, $d=[0.3,0.5,1.0]$ (in meters). For each distance, hand gestures were picturing with two different scales, $r=[0.2,0.5]$ (in meters). In our experiments, we capture all the samples with 4 volunteers, for each gestures of certain distance and scale. In total, 38400 hand gesture samples were completely collected to the database.

\subsection{Time-Frequency Analysis}

In this subsection, we apply time-frequency analysis to see more feature difference between 24 reference gestures. Time-frequency analysis is good way to do that, which is a two-dimensional figure picturing the signal's spectrum change with time. The most common methods of time-frequency analysis are short-time Fourier transform (STFT)\cite{sejdic2009time} and continuous wavelet transform (CWT)\cite{mallat1999wavelet}.z

The short-time Fourier transform or the short-term Fourier transform is a Fourier-dependent transformation for determining the sinusoidal frequency and the phase content of the local portion of the signal over time. Compared with the Fourier transform, it focuses on instantaneous frequency information.

But STFT has a small drawback, that is, the width of the time window is inversely proportional to the width of the frequency window. This means that high resolution in the time domain results in low resolution being the frequency domain and vice versa. The width of the sliding window limits the frequency resolution of the STFT. Continuous wavelet transform (CWT) is another wonderful way to solve this problem.

The result of the time-frequency analysis are the foundations of the following classification.

\section{Advanced Classification based on Deformable Deep Convolutional Generative Adversarial Networks}
Based on the features extracted from the last section, the next step in to use apply classification to justify those defined gestures.

In the original design, we applied original convolutional neural networks, which is really effective on our samples, as classification method. However, with the experiments going on, limits on the original convolutional neural network architecture came up. The most obvious ones are as follows:

\begin{itemize}
	\item With the expanding class of gesture classes, the over-fitting effect of the classification increases due to the limited sample of hand gestures. Since all of the samples are retrieved by real people collection and the increasing demand of hand gesture samples became a burden, the classification method need improvement to eliminate the false effect from the size of database.
	\item During the real-life testing, the processing time of testing is long if the hardware is aged or outdated. The main reason of the long testing time is data redundancy. We should find a way to train the network to recognize the unnecessary information during the training process to shorten the time of testing.
\end{itemize}

Focusing on these two limitation, we are ready to tackle the problem one by one combining based on Deformable Deep Convolutional Generative Adversarial Networks(De-DCGAN), which is an improved version of DCGAN.

Core to our approach to De-DCGAN is adopting and modifying two recently demonstrated changes to CNN architectures.

To begin with, background of Generative adversarial networks(GANs)\cite{goodfellow2014generative} needs to be explained. Generated adversarial networks is an artificial intelligence algorithm for unsupervised machine learning. It is implemented by the system of two neural networks competing with each other in the zero-sum game framework. They were first introduced by Ian Goodfellow et al. The main component of GAN is two competing networks, a generating model $ G $ capture data distribution, and a discriminant model $ D $, which estimates the probability that the sample comes from training data rather than $G$.\cite{luc2016semantic}

Deep Convolutional Generative Adversarial Networks(DCGAN)\cite{radford2015unsupervised} is an improved convolutional networks. The main contribution of DCGAN team is they propose and evaluate a set of constraints on the architectural topology of Convolutional GANs that make them stable to train in most settings. By far, it is the most stable architecture to train in most settings and the most convenient architecture for both generating samples and improving classification rate.

Another modification is Deformable Convolutional Networks(DCN)\cite{dai2017deformable}. The traditional convolutional neural networks are inherently limited to model geometric transformations due to the ﬁxed geometric structures in its building modules. DCN introduces two new modules to enhance the transformation modeling capacity of CNNs, namely, deformable convolution and deformable RoI pooling. The new modules can readily replace their plain counterparts in existing CNNs and can be easily trained end-to-end by standard back-propagation, giving rise to deformable convolutional networks.

\begin{figure*}
	\centerline{\includegraphics[height=6cm,clip]{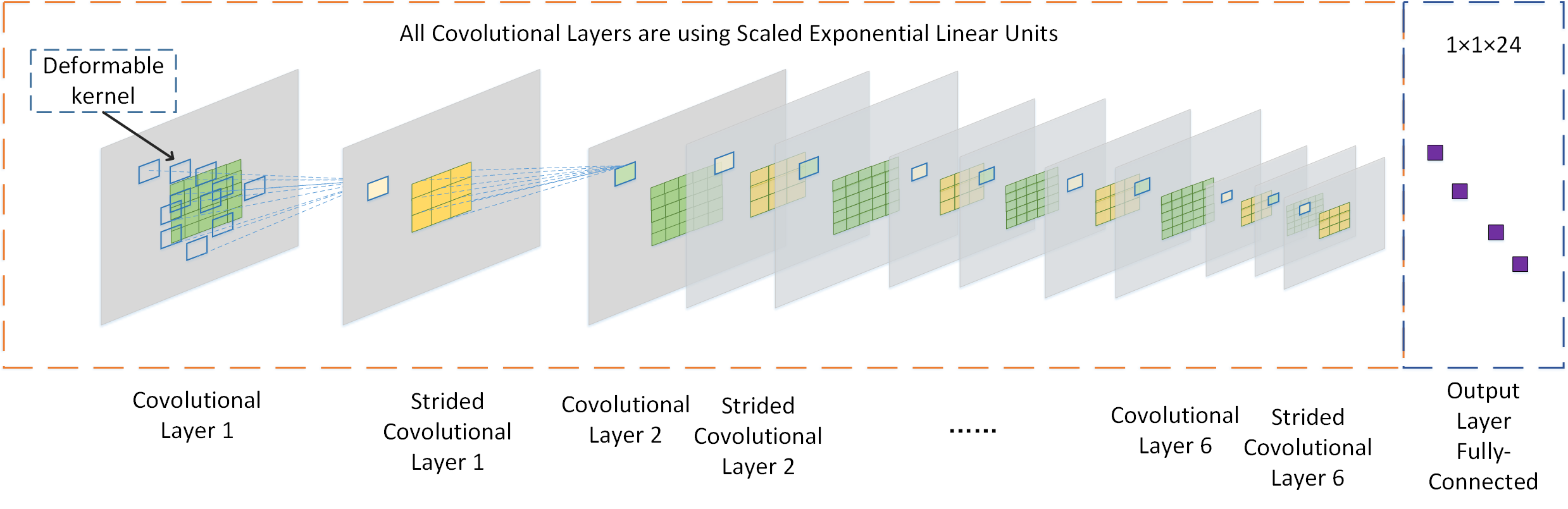}}
	\caption{De-DCGAN Architecture}
	\label{De-DCGAN} 		
\end{figure*}

We propose De-DCGAN based on the two methods above and create a new architecture in our scenario as in Fig.\ref{De-DCGAN}. 

De-DCGAN firstly did the following modifications based on the original CNN architecture.

\subsection{Replace pooling layers with strided convolutions in discriminator and fractional-strided convolutions in generators}

Pooling layers are commonly used for conventional convolution neural networks. In order to understand the effect of this alternative, we must review the traditional CNN architecture of the convolution layer and layer function. We denote the feature map produced by certain layer of a CNN by $f$. It can be taken as a 3-dimensional matrix of size $W\times H\times N$ where $W$ and $H$ means the width and height and the $N$ is the number of channels. Then p-norm pooling with size $k$ and stride $s$ applied to the featured map $f$ is a featured map $f$ is a 3D matrix $s(f)$ with the following entries:
\begin{align}
s_{i,j,u}(f)=(\sum_{h=-k}^{k} \sum_{w=-k}^{k} |f_{g(h,w,i,j,u)}|^p)^{1/p}
\end{align}
where $g(h,w,i,j,u)=s\cdot i+h,s\cdot j + w,u)$ is the translation to calculate the according position in $s$ to positions in $f$ with respect to $s$, $p$ is the norm order of $p-norm$ (when $p$ comes to infinite, it equals max-pooling layer).Usually, pooling layers don't overlap, which occurs when $r > k$; but the conventional CNN architecture we used includes overlap pooling layers with $k = 3 $ and $r = 2$. Compare the pooling layer defined above to the convolutional layer applied to the feature defined as:

\begin{align}
c_{i,j,o}(f)=\lambda(\sum_{h=-k}^{k} \sum_{w=-k}^{k} \sum_{u=1}^{N} \theta_{h,w,u,o}\cdot f_{g(h,w,u,o)})
\end{align}
where $\theta$ are the local weights (or the kernel weights, or filters), $\lambda(\cdot)$ is the activation function, which will be discussed later, and $o \in [1, M]$ is the number of output feature (or channel) of the convolutional layer.

After comparing we noticed that the main function of the pooling layers are the following two: 1) reduction of the spatial dimension which can enlarge the input size possibility 2)the p-norm makes the representation in a CNN more invariant. Since only the second is the crucial part of a CNN architecture, we try to replace the pooling layer with a convolutional layer with certain stride. For example, to replace a pooling layer with a $3\times3$ pooling kernel and stride parameter $r=2$, we replace it with a convolutional layer with same stride and corresponding kernel size and same number of output channels.

\subsection{Use Scaled Exponential Linear Units in Generators and Discriminators}

The second modification is that we use a novel activation layer called "scaled exponential linear units"(SELU) in our networks, both in generators and discriminators.

In traditional CNN networks, people use rectified linear unit(ReLU) as activation layer. It has became the most commonly used activation layer since it has been brought up in 2012\cite{krizhevsky2012imagenet} .Compared to tanh and sigmoid, it has great qualities over like non-saturating and quick convergence speed. However, it has its own limit. ReLU is highly dependent to initialization and it can easy cause dead neurons if the learning rate is too big. That's where batch normalization(BN) comes\cite{ioffe2015batch}. Batch Normalization is to normalize the ReLU layer at each batch, which maintains the zero mean and unit variance of the activation output. By applying batch Normalization, we can not only improve the gradient flow through the network, but also allows higher learning rates. Besides, it can reduce the strong dependence on initialization. However, the application of BN to each activation layer increase the complexity and training time of each epoch.

Recently, a new activation layer called "scaled exponential linear units"(SELU) is proposed\cite{klambauer2017self}. SELU has all the best properties that ReLU has. Besides, the best part of SELU activations is that it can guarantee the output of each of many network layers will converge towards zero mean and unit variance. So we can discard Batch Normalization layer after every activation layer and ease the burden of longer training time caused by them. Due to its great properties, we adopt SELU in generators and discriminators. The SELU activation function is given by 

\begin{align}
\text{selu}(x)=\lambda \begin{cases}
x& \text{if }x>0\\
\alpha e^x-\alpha& \text{if } x\leq 0
\end{cases}
\end{align}

The application of SELU in our architectures allows us to train it in deep network with many layer. Besides, it employs strong regularization schemes, realize high learning and achieves high stability of the model. 

In order to compare the performance of SELU, we will show you a comparison test between BN+ReLU and SELU in the next section.

\subsection{Use Deformable Convolutional Kernel}

Besides, we transform all the convolutional layer into deformable ones. A traditional 2D convolution comes the two steps: 1) Use the regular grid $R$ to sample the input feature graph $x$; 2) The sum of the sampled values weighted by $w$. The grid $ R $ defines the size and expansion of the receiving field. For example, $R={(-1,1),(-1,0),...,(0,1),(1,1)}$ deﬁnes a 5$\times$5 kernel with dilation 1.

For the output function graph $y$ for each location $p_0$, we have 
\begin{align}
y(p_0)=\sum_{P_n\in R}^{ } w(p_n)*x(p_0+p_n)
\end{align}

The deformable convolution increases the conventional grid $R$ with the offset $\{\bigtriangleup p_n |n = 1, ..., N\}$, where $N = |R|$. Eq.(4) becomes

\begin{align}
y(p_0)=\sum_{P_n\in R}^{ } w(p_n)*x(p_0+p_n+\bigtriangleup p_n)
\end{align}

Now, the sampling is over the irregular and offset locations $p_n +\bigtriangleup p_n$ . As the offset $\bigtriangleup p$ n is typically fractional, Eq. (2) is implemented via bilinear interpolation as

\begin{align}
x(p)=\sum_{q}^{ } G(q,p)\cdot x(q)
\end{align}

where $p$ denotes an arbitrary (fractional) location ($p = p_0+p_n+\bigtriangleup p_n$ for Eq. (2)), q enumerates all integral spatial locations in the feature map $x$, and $G(\cdot, \cdot)$ is the bilinear interpolation kernel. Note that $G$ is two dimensional. It is separated into two one dimensional kernels as

\begin{align}
G(q,p)=g(q_x,p_x)\cdot g(q_y,p_y)
\end{align}

where $g(a, b) = max(0, 1-|a-b|)$. Eq. (3) is fast to compute as conv $G(q, p)$ is non-zero only for a few $q$s.

As shown in Fig. 1, the offset is obtained by applying a convolution layer on the same input feature map. The convolution kernel has the same spatial resolution as the current convolution layer (for example, $ 5 \ times 5 $ in Figure 4). It is obvious that the output offset field has the same spatial resolution as the input feature map. 

Encoding the $ N2D $ offset vector, the channel dimension is $2N$. During the training process, you can understand the convolution kernel used to generate the output feature and generate the offset. The gradient that is enforced on the deformable convolution module can be propagated backwards by the bilinear mapping operation in the formula. (6) and (7).

\begin{figure}[htb]
	\centerline{\includegraphics[height=4.5cm,clip]{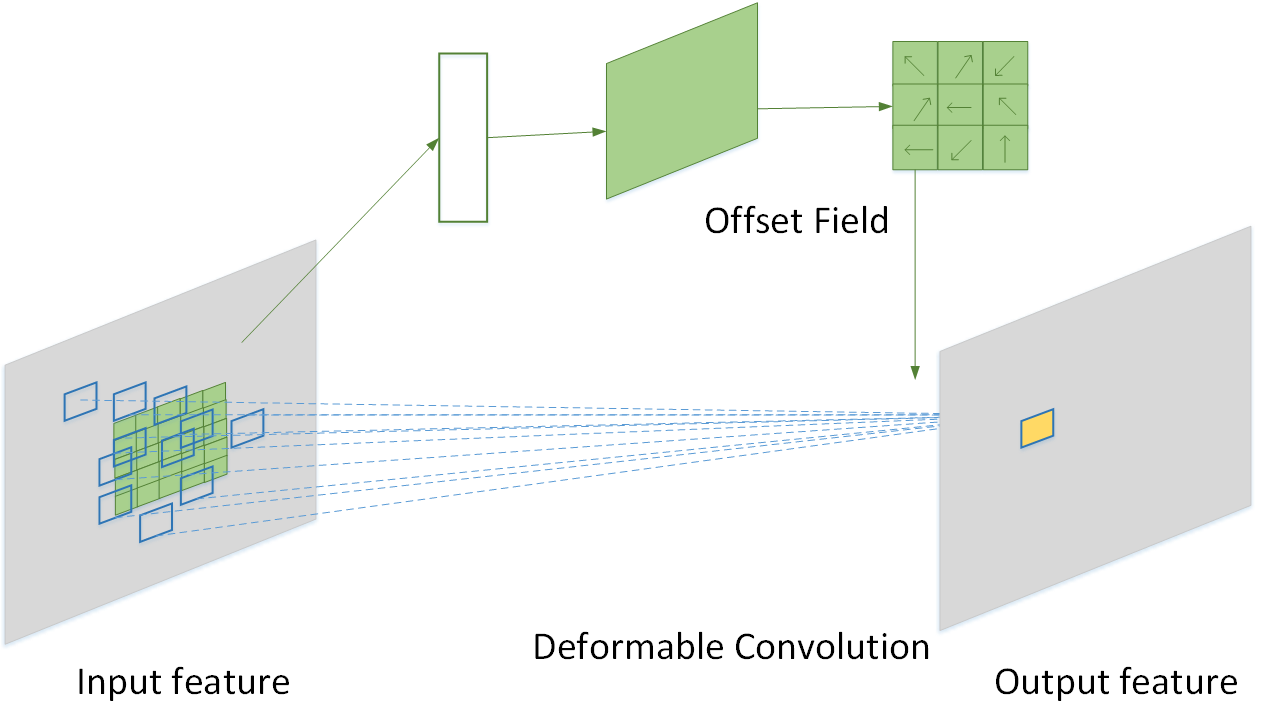}}
	\caption{Illustration of 5 $\times$ 5 deformable convolution.}
	\label{Illustration}
\end{figure}

\section{Experiment Result}

Since this design mainly focus on the new classification architecture, we mainly focuses on four questions:
\begin{itemize}
	\item Whether De-DCGAN architecture can generate samples with good quality
	\item Whether De-DCGAN architecture can reduce the over-fitting and increase the recognition rate
	\item Whether SELU activations improve the architecture's performance
	\item Whether Deformable kernel can eliminate the testing time cost
\end{itemize}

We trained our samples on the NVEDIA TITAN GPU and test the result using MacBook 2016 CPU.

\begin{figure}[htb]
	\centering{
		\includegraphics[height=5.5cm,width=6cm]{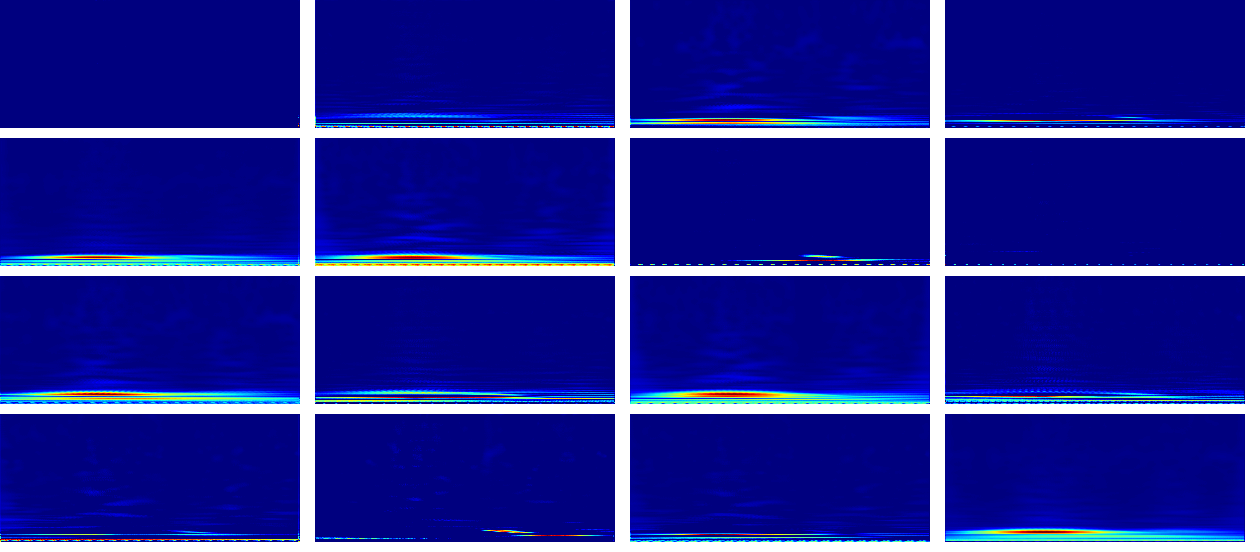}}
	\caption{Generated pictures}
	\label{Generatedpictures}		
\end{figure}

For the first question, after 25 epochs, we can see from that the training process can give out the good quality generated pictures as Fig.\ref{Generatedpictures} to enlarge the database for classification.

For the second question, we cross examined the recognition testing rate using four original (real data, not generated ones) on two independent trained model, one based on the traditional CNN architecture, another one based on the new De-DCGAN architecture. Both models are trained by 50 epochs. The result shows in the Table 1.

\begin{table}[htbp]
	\centering 
	\caption{Recognition Rate with traditional CNN and De-DCGAN} 
	\begin{tabular}{lccc}  
		\hline
		Average Rate & Traditional CNN & De-DCGAN\\ \hline  % \hline 在此行下面画一横线
		Gesture 1 &72.3\% &87.4\%\\         % \\ 表示重新开始一行
		Gesture 2 &79.8\% &93.3\%\\        % & 表示列的分隔线
		Gesture 3 &75.7\% &85.7\%\\
		Gesture 4 &76.7\% &90.9\%\\ \hline
	\end{tabular}
\end{table}

For the third question, to test the performance and quality of SELU activation layer, we bring three sets as comparisons on the same dataset: ReLU+BN, SELU+BN and just SELU. The second corresponding set is to prove the SELU activation has the same good quality as the combination of ReLU and BN.

\begin{table}[htbp]
	\centering 
	\caption{Activations Layer Comparison: BN+ReLU or SELU} 
	\begin{tabular}{lccc}  
		\hline
		Average Rate & ReLU+BN &SELU+BN &SELU\\ \hline  % \hline 在此行下面画一横线
		Gesture 1 &79.5\% &89.6\% &87.6\%\\         % \\ 表示重新开始一行
		Gesture 2 &78.2\% &88.7\% &90.6\%\\        % & 表示列的分隔线
		Gesture 3 &77.6\% &85.7\% &85.7\%\\
		Gesture 4 &74.9\% &80.9\% &81.0\%\\ \hline
	\end{tabular}
\end{table}

As we can see from the test result, the testing result tells us that SELU has a better quality than ReLU from the testing recognition rate. And the rate between SELU and SELU+BN proves that SELU itself has the convergence quality of BN. Thus the rate did not change that much after cutting down the BN layer.

Lastly, for the last question, we examined the recognition time with the traditional architecture with normal convolution kernel and the De-DCGAN with the deformable kernel. The result shows in the Table 3.

\begin{table}[htbp]
	\centering 
	\caption{Testing time with traditional CNN and De-DCGAN} 
	\begin{tabular}{lccc}  
		\hline
		Testing time & Traditional CNN & De-DCGAN\\ \hline  % \hline 在此行下面画一横线
		Gesture 1 & 590ms & 360ms\\         % \\ 表示重新开始一行
		Gesture 2 & 580ms & 366ms\\        % & 表示列的分隔线
		Gesture 3 & 700ms & 479ms\\
		Gesture 4 & 706ms & 482ms\\ \hline
	\end{tabular}
\end{table}

As we can see from the Table 2, the testing time decreases like 30\% in average. Actually, the time cost decrease comes from the increase of the training time. But since we don't have constraints on the training time in our case, the result is acceptable.

\section{Conclusion}
This paper proposes a Deformable Deep Convolutional Generated Adversarial Networks used on hand gesture recognition systems. Our system adopted Doppler-radar sensor with dual receiving channels at 5.8 GHz to acquire large hand gesture reference library of 24 hand gesture combinations. We applied two time-frequency analysis, short-time Fourier transform and continuous wavelet transform, as the foundation of the classification. The main focus of the work is to tackle two problems rising with the expanding of gestures and limited database:the over-fitting performance and the long testing time. 

The major contribution of our work is we proposed a Deformable Deep Convolutional Generated Adversarial Networks as the classification method to solve these two problems. At first, we did some modifications to the original GAN architecture to suit our scenario. Besides, we adopted deformable convolutional kernel in all the convolutional layers. The experiments shows that the proposed architecture not only increase the correct recognition rate with different gesture combinations, but also decrease the time of testing. 

To sum up, our proposed system proves the feasibility of the De-DCGAN of tackling time and over-fitting problems. And the experiments shows a great potential on De-DCGAN's performance on radar-based gesture recognition system.

\section*{Acknowledgments}
This work is supported by Intel under agreement No. CG\# 30397855 and the Fundamental Research Funds for the Central Universities (No.2017QNA5009).

\bibliographystyle{IEEEtran}
\bibliography{WCSP}

\begin{thebibliography}{10}
\providecommand{\url}[1]{#1}
\csname url@samestyle\endcsname
\providecommand{\newblock}{\relax}
\providecommand{\bibinfo}[2]{#2}
\providecommand{\BIBentrySTDinterwordspacing}{\spaceskip=0pt\relax}
\providecommand{\BIBentryALTinterwordstretchfactor}{4}
\providecommand{\BIBentryALTinterwordspacing}{\spaceskip=\fontdimen2\font plus
\BIBentryALTinterwordstretchfactor\fontdimen3\font minus
  \fontdimen4\font\relax}
\providecommand{\BIBforeignlanguage}[2]{{%
\expandafter\ifx\csname l@#1\endcsname\relax
\typeout{** WARNING: IEEEtran.bst: No hyphenation pattern has been}%
\typeout{** loaded for the language `#1'. Using the pattern for}%
\typeout{** the default language instead.}%
\else
\language=\csname l@#1\endcsname
\fi
#2}}
\providecommand{\BIBdecl}{\relax}
\BIBdecl

\bibitem{hjelmaas2001face}
E.~Hjelm{\aa}s and B.~K. Low, ``Face detection: A survey,'' \emph{Computer
  vision and image understanding}, vol.~83, no.~3, pp. 236--274, 2001.

\bibitem{zhao2003face}
W.~Zhao, R.~Chellappa, P.~J. Phillips, and A.~Rosenfeld, ``Face recognition: A
  literature survey,'' \emph{ACM computing surveys (CSUR)}, vol.~35, no.~4, pp.
  399--458, 2003.

\bibitem{molchanov2015multi}
P.~Molchanov, S.~Gupta, K.~Kim, and K.~Pulli, ``Multi-sensor system for
  driver's hand-gesture recognition,'' in \emph{Automatic Face and Gesture
  Recognition (FG), 2015 11th IEEE International Conference and Workshops on},
  vol.~1.\hskip 1em plus 0.5em minus 0.4em\relax IEEE, 2015, pp. 1--8.

\bibitem{molchanov2015short}
------, ``Short-range fmcw monopulse radar for hand-gesture sensing,'' in
  \emph{RadarCon}.\hskip 1em plus 0.5em minus 0.4em\relax IEEE, 2015, pp.
  1491--1496.

\bibitem{arbabian201394}
A.~Arbabian, S.~Callender, S.~Kang, M.~Rangwala, and A.~M. Niknejad, ``A 94 ghz
  mm-wave-to-baseband pulsed-radar transceiver with applications in imaging and
  gesture recognition,'' \emph{IEEE Journal of Solid-State Circuits}, vol.~48,
  no.~4, pp. 1055--1071, 2013.

\bibitem{hugler2016rcs}
P.~H{\"u}gler, M.~Geiger, and C.~Waldschmidt, ``Rcs measurements of a human
  hand for radar-based gesture recognition at e-band,'' in \emph{Microwave
  Conference (GeMiC), 2016 German}.\hskip 1em plus 0.5em minus 0.4em\relax
  IEEE, 2016, pp. 259--262.

\bibitem{pu2013whole}
Q.~Pu, S.~Gupta, S.~Gollakota, and S.~Patel, ``Whole-home gesture recognition
  using wireless signals,'' in \emph{Proceedings of the 19th annual
  international conference on Mobile computing \& networking}.\hskip 1em plus
  0.5em minus 0.4em\relax ACM, 2013, pp. 27--38.

\bibitem{mercuri2013analysis}
M.~Mercuri, P.~J. Soh, G.~Pandey, P.~Karsmakers, G.~A. Vandenbosch, P.~Leroux,
  and D.~Schreurs, ``Analysis of an indoor biomedical radar-based system for
  health monitoring,'' \emph{IEEE Transactions on Microwave Theory and
  Techniques}, vol.~61, no.~5, pp. 2061--2068, 2013.

\bibitem{zhou2016ultra}
Z.~Zhou, J.~Zhang, and Y.~D. Zhang, ``Ultra-wideband radar and vision based
  human motion classification for assisted living,'' in \emph{Sensor Array and
  Multichannel Signal Processing Workshop (SAM), 2016 IEEE}.\hskip 1em plus
  0.5em minus 0.4em\relax IEEE, 2016, pp. 1--5.

\bibitem{peng2016fmcw}
Z.~Peng, J.-M. Mu{\~n}oz-Ferreras, R.~G{\'o}mez-Garc{\'\i}a, and C.~Li, ``Fmcw
  radar fall detection based on isar processing utilizing the properties of
  rcs, range, and doppler,'' in \emph{Microwave Symposium (IMS), 2016 IEEE
  MTT-S International}.\hskip 1em plus 0.5em minus 0.4em\relax IEEE, 2016, pp.
  1--3.

\bibitem{sejdic2009time}
E.~Sejdi{\'c}, I.~Djurovi{\'c}, and J.~Jiang, ``Time--frequency feature
  representation using energy concentration: An overview of recent advances,''
  \emph{Digital Signal Processing}, vol.~19, no.~1, pp. 153--183, 2009.

\bibitem{mallat1999wavelet}
S.~Mallat, \emph{A wavelet tour of signal processing}.\hskip 1em plus 0.5em
  minus 0.4em\relax Academic press, 1999.

\bibitem{goodfellow2014generative}
I.~Goodfellow, J.~Pouget-Abadie, M.~Mirza, B.~Xu, D.~Warde-Farley, S.~Ozair,
  A.~Courville, and Y.~Bengio, ``Generative adversarial nets,'' in
  \emph{Advances in neural information processing systems}, 2014, pp.
  2672--2680.

\bibitem{luc2016semantic}
P.~Luc, C.~Couprie, S.~Chintala, and J.~Verbeek, ``Semantic segmentation using
  adversarial networks,'' \emph{arXiv preprint arXiv:1611.08408}, 2016.

\bibitem{radford2015unsupervised}
A.~Radford, L.~Metz, and S.~Chintala, ``Unsupervised representation learning
  with deep convolutional generative adversarial networks,'' \emph{arXiv
  preprint arXiv:1511.06434}, 2015.

\bibitem{dai2017deformable}
J.~Dai, H.~Qi, Y.~Xiong, Y.~Li, G.~Zhang, H.~Hu, and Y.~Wei, ``Deformable
  convolutional networks,'' \emph{arXiv preprint arXiv:1703.06211}, 2017.

\bibitem{krizhevsky2012imagenet}
A.~Krizhevsky, I.~Sutskever, and G.~E. Hinton, ``Imagenet classification with
  deep convolutional neural networks,'' in \emph{Advances in neural information
  processing systems}, 2012, pp. 1097--1105.

\bibitem{ioffe2015batch}
S.~Ioffe and C.~Szegedy, ``Batch normalization: Accelerating deep network
  training by reducing internal covariate shift,'' in \emph{International
  Conference on Machine Learning}, 2015, pp. 448--456.

\bibitem{klambauer2017self}
G.~Klambauer, T.~Unterthiner, A.~Mayr, and S.~Hochreiter, ``Self-normalizing
  neural networks,'' \emph{arXiv preprint arXiv:1706.02515}, 2017.

\end{thebibliography}

\end{document}